\documentclass{edm_article}
\usepackage{graphicx}      % For adding figures
\usepackage{booktabs}      % For beautiful tables
\usepackage{geometry}      % For page margins (optional, if using conference template)
\usepackage{enumitem}      % For compact lists [noitemsep]
\usepackage{amsmath}       % For math equations
\usepackage{microtype}     % VITAL: Prevents text overflow into margins!
\usepackage{url}

\usepackage{soul}
\usepackage{booktabs}
\usepackage{xcolor}
\usepackage{float}
\title{ %Classifying LLM Reasoning to Validate Large-Scale Educational Coding
   LLM Reasoning Predicts When Models Are Right: Evidence from Coding Classroom Discourse
}

\numberofauthors{5}
\author{
Bakhtawar Ahtisham\\
\affaddr{Cornell University}\\
\email{ba453@cornell.edu}
\and
Kirk Vanacore\\
\affaddr{Cornell University}\\
\email{kpv27@cornell.edu}
\and
Zhuqian Zhou\\
\affaddr{Cornell University}\\
\email{zz968@cornell.edu}
\and
Jinsook Lee\\
\affaddr{Cornell University}\\
\email{jl3369@cornell.edu}
\and
Rene F. Kizilcec\\
\affaddr{Cornell University}\\
\email{kizilcec@cornell.edu}
}

\begin{document}

\maketitle

\begin{abstract}
Large Language Models (LLMs) are increasingly deployed to automatically label and analyze educational dialogue at scale, yet current pipelines lack reliable ways to detect when the models are wrong. We investigate whether reasoning generated by LLMs can be used to predict the correctness of a model's own predictions. We analyzed 30,300 teacher utterances from classroom dialogue, each labeled by multiple state-of-the-art LLMs with an instructional move construct and an accompanying reasoning. Using human-verified ground-truth labels, we framed the task as predicting whether a model's assigned label for a given utterance was correct. We encoded the LLM reasoning using Term Frequency-Inverse Document Frequency (TF-IDF) and evaluated five supervised classifiers on their ability to predict label correctness. A Random Forest classifier achieved an F1 score of 0.83 (Recall = 0.854), successfully identifying the majority of incorrect predictions and significantly outperforming baselines. Training specialist detectors for specific instructional move constructs further improved performance on difficult constructs, suggesting that error detection benefits from construct-specific linguistic cues. Using the Linguistic Inquiry and Word Count (LIWC) framework, we examined four linguistic constructs as markers of correctness: Causation (logical reasoning), Differentiation (complexity), Tentativeness (uncertainty), and Insight (metacognition). Correct predictions were characterized by grounded causal logic (e.g., `because', `therefore'), while faulty reasoning was five times more likely to rely on epistemic hedging (e.g., `might', `could') and twice as likely to employ performative metacognition (e.g., `think', `realize'). Syntactic complexity did not distinguish correct from incorrect reasoning, indicating that evidential grounding --- not structural sophistication --- drives validity. Our findings also challenge the ``longer reasoning is better" heuristic; for most constructs, concise reasoning is a stronger predictor of correctness. These findings suggest that reasoning-based error detection provides a practical and scalable approach to quality control in automated educational dialogue analysis. 

\end{abstract}

 % Replace with your own 3-5 keywords

\section{Introduction}
The analysis of qualitative educational data is increasingly central to Educational Data Mining (EDM) and related fields \cite{d2010mining,zambrano2026data}. Researchers routinely annotate discussion posts, assignments, or dialogue with specific constructs to support downstream analyses, such as Epistemic Network Analysis (ENA) \cite{cai2017epistemic} or sequential pattern mining \cite{vanacore2025downshifting}. These annotations enable large-scale investigations of how students and educators engage in learning tasks in classrooms \cite{petrilli2024next}, tutoring sessions \cite{demszky2023mpowering}, and online chat environments. Historically, this work has relied on supervised classification methods \cite{cai2019ncoder,demszky2021measuring}, and more recently, Large Language Models (LLMs) to annotate data at scale \cite{Zambrano_Liu_Barany_Baker_Kim_Nasiar_2023,Liu_2025}. However, automated annotation pipelines often exhibit low to moderate reliability and inconsistent generalization across constructs \cite{zarisheva2024deductive,kim2025code}. In particular, false positives remain a persistent challenge because they can systematically bias downstream structural and sequential modeling \cite{vanacore2025well}.

This paper addresses this central reliability gap: current pipelines lack robust mechanisms for detecting when automated annotations are wrong. Existing work primarily focuses on improving model accuracy, yet even high-performing models produce errors that propagate into subsequent analyses. We propose a complementary approach: using the reasoning generated by LLMs as a diagnostic signal for identifying misclassifications. Rather than treating LLMs as black-box classifiers, we examine whether their reasoning can serve as a secondary layer of quality control. 

We operationalize this idea by training models to predict whether an LLM-generated annotation is correct or a false positive based solely on the accompanying reasoning text--asking the LLM to provide reasoning along with the annotation. Using traditional natural language processing (NLP) and machine learning (ML) methods, we show that correct and incorrect annotations can be distinguished with high accuracy ($F_1$ = 0.830 to 0.854), with further improvements when models are trained for specific instructional constructs (e.g., separate models for classifying ``Press for Accuracy'' and ``Press for Reasoning''). We additionally analyze linguistic markers associated with misclassification, finding that reasoning supporting false positives is longer and contains substantially more epistemic hedging, suggesting that models often verbalize their own uncertainty. Together, these findings introduce reasoning-based error detection as a scalable strategy for improving the robustness of automated discourse analysis in education.

This paper makes three main contributions: First, we introduce reasoning-based error detection as a new reliability layer for automated educational discourse annotation. Second, we demonstrate that traditional NLP and ML methods can accurately predict LLM annotation misclassification using only the LLM reasoning. And third, we identify linguistic markers of correctness and error in LLM reasoning, providing new insight into how and when automated annotations fail.

% 1. Main Finding: Can we do it? (
% 2. The "Why" (Linguistics): Why does it work? (Hedge words = Uncertainty, Length = Waffling).
% 3. Robustness: Does it work everywhere? (Yes, across all 4 LLMs).
% 4. Optimization: Can we make it better? (Specialist models).

\section{Related Work}

\subsection{The Shift to Large Language Models in Classification}
The EDM community has a long history of using machine learning and AI methods for classifying complex data at scale, ranging from students' knowledge states \cite{corbett1994knowledge,piech2015deep} to epistemic affect \cite{dmello2012dynamics,pardos2011affect}. While much of this classification has dealt with structured log-file data using supervised machine learning models, recent work has shifted towards classification using Large Language Models (LLMs). This shift has been particularly beneficial for processing unstructured data, such as open-ended math problems \cite{siedahmed2025nonstandard}, dialogues in tutoring sessions \cite{acosta2025recognizing,zhang2024detecting}, classroom discourse \cite{tran2024analyzing,whitehill2023automated}, and writing tasks \cite{neshaei2025bridging,latif2024artificial}.

Unlike many classification methods that require extensive feature engineering or large labeled datasets, LLMs offer the potential for ``zero-shot'' or ``few-shot'' classification, allowing researchers to scale qualitative coding to datasets previously considered too costly to annotate manually \cite{kim2025code,gilardi2023chatgpt,ziems2023can}. To improve classification accuracy, researchers have employed a variety of methods beyond simple prompting. For instance, recent studies have demonstrated that including diverse few-shot examples significantly boosts performance in identifying collaborative argumentation moves \cite{tran2025collaborative} and assessing instructional quality \cite{petrilli2024next}. Furthermore, the integration of reasoning—asking the model to ``think'' before it classifies—has been shown to improve the detection of latent constructs like self-regulated learning processes \cite{zhang2024detecting}. By explicitly modeling the inference steps, these methods aim to bridge the gap between surface-level keywords and the deep semantic understanding required for qualitative coding.

\subsection{Challenges with Misclassifications, Faithfulness}
Despite these methodological advances, the reliability of LLM-based annotation remains inconsistent. While some studies report high agreement between LLMs and human experts for broad categories \cite{neshaei2025bridging}, others highlight significant declines in performance when models encounter ambiguous or high-inference constructs \cite{kim2025code}. Crucially, \cite{zarisheva2024deductive} found that LLM precision in deductive coding tasks can vary drastically (from 0.10 to 0.87) depending on the specific prompt and construct. Similarly, \cite{vanacore2025well} observed that LLMs consistently exhibit a high false positive rate, even when achieving moderate reliability overall.

This phenomenon is often described in NLP literature as a ``faithfulness failure''—where the model generates a label that is plausible on the surface but contradicts the deeper semantic meaning of the input text \cite{ji2023survey,agarwal2024faithfulness,jacovi2020faithfulness,madsen2024selfexplanations}. In educational contexts, this often manifests as the model over-relying on specific keywords (e.g., classifying any question starting with ``Why'' as a \textit{Press for Reasoning}) rather than analyzing the true pedagogical intent \cite{vanacore2025well}. 

\subsection{LLM Reasoning as Diagnostic Signals}
Recent literature suggests that LLMs possess an internal awareness of their own correctness, which can be elicited through probability estimation \cite{kadavath2022language} or verbalized confidence scores \cite{tian2023just,jiang2021calibration,lin2022confidence}. However, explicit confidence prompts do not always capture the nuance of model failure, and existing mitigation strategies often rely on sampling consistency. Methods such as \textit{SelfCheckGPT} \cite{manakul2023selfcheckgpt} or \textit{Self-Consistency} \cite{wang2022self} generate multiple responses for the same input and aggregate them to filter out outliers. While effective, these methods increase computational cost linearly with the number of samples ($N$), making them prohibitively expensive for real-time educational feedback.

Consequently, a growing body of work suggests that the structure and content of model-generated explanations can serve as diagnostic signals for identifying incorrect predictions. In prior literature, this signal is often studied through reasoning traces—step-by-step intermediate reasoning sequences produced during inference, commonly in chain-of-thought or process supervision settings. While \cite{turpin2023language} caution that such traces may function as unfaithful post-hoc rationalizations, recent work in process supervision has nevertheless shown that properties of these traces, such as length and linguistic hedging, correlate with model uncertainty and error \cite{xiong2025trace, tao2025uncertainty, su2025underthinking}. In contrast, the explanations analyzed in this study are elicited rationales: short, post-hoc justifications generated alongside a predicted label rather than procedural traces of internal inference. We do not assume these rationales faithfully reflect latent reasoning processes. Instead, we treat them as observable linguistic artifacts whose structure may still provide diagnostic information. Building on insights from trace-based work, we operationalize these linguistic markers within the educational domain to filter false positives in classroom discourse analysis, without the computational overhead of multiple sampling or process-level supervision.

% Consequently, a growing body of work indicates that the structure and content of a \textit{single} reasoning trace can serve as a diagnostic signal. While \cite{turpin2023language} warned that LLM reasoning can be unfaithful post-hoc rationalizations, recent studies in process supervision have leveraged these traces to detect errors. Most notably, \cite{xiong2025trace} identified a strong correlation between reasoning length and uncertainty, suggesting that models generate more verbose and hedged explanations when attempting to justify incorrect classifications. Our work operationalizes these findings within the educational domain, using these linguistic markers to filter false positives in classroom discourse analysis without the computational overhead of multiple sampling.

\subsection{Linguistic Analysis of Generated Text}
To understand the nature of these failures, researchers have turned to linguistic analysis tools like Linguistic Inquiry and Word Count (LIWC) \cite{pennebaker2015development}. LIWC counts words in psychologically meaningful categories (e.g., affect, cognition, certainty) and has been widely used to analyze student essays, forum posts, and increasingly, AI-generated text. Prior work evaluating human-AI differences suggests that AI-generated text often exhibits distinct linguistic profiles, such as lower variance in vocabulary and specific patterns of hedging when the model is "confabulating" or hallucinating \cite{yao2024editing}.
Similarly, \cite{xiong2025trace} identified that uncertainty in reasoning is often marked by increased verbosity and "hedging" language. Our work extends this methodology to the validation of automated coding. By applying LIWC analysis to LLM reasoning, we isolate the specific linguistic markers—such as high "insight" and low "causation"—that differentiate valid, grounded classification from misclassified outputs.

\section{The Current Study}
The increasing use of Large Language Models (LLMs) for automated discourse annotation has enabled educational researchers to analyze classroom dialogue at unprecedented scale. However, despite improvements in classification accuracy, LLM-based annotation pipelines continue to suffer from persistent misclassifications—particularly false positives—that can systematically bias downstream analyses such as sequential modeling and epistemic network analysis. Existing work has largely focused on improving model performance, with comparatively less attention paid to mechanisms for detecting when an automated annotation is likely incorrect.

The current study investigates whether model-generated explanations—elicited alongside predicted instructional move labels—can serve as a diagnostic signal for identifying annotation errors. Rather than treating LLMs as black-box classifiers, we examine whether the linguistic structure of the explanations they produce when justifying a label provides useful information about the correctness of that label. This framing positions explanation-based verification as a complementary reliability layer that operates independently of the original classification model and does not require additional sampling or access to model internals. We evaluate this approach in the context of automated coding of classroom discourse using a well-established Talk Moves framework. Specifically, we treat misclassification detection as a secondary supervised learning task: given an LLM’s predicted label and its accompanying explanation, can we predict whether the label is correct relative to expert human annotations? In addition to assessing whether such a signal exists, we examine how robust it is across different state-of-the-art LLM architectures, whether it can be explained in terms of interpretable linguistic features, and whether performance improves when verification models are specialized for individual instructional constructs.

Guided by this framing, the study addresses the following research questions:

% The current study evaluates how LLM reasoning can be used to identify annotation errors—``misscasslifciations''—and evaluate linguistic markers of these errors by addressing the following questions: 

\textbf{RQ1:} Can the reasoning generated by LLMs serve as a reliable diagnostic signal for detecting misclassifications in educational discourse analysis?\\
\textbf{RQ2:} How robust is this reasoning-based verification signal across different state-of-the-art LLM architectures?\\
\textbf{RQ3:} What specific linguistic features within LLM reasoning are most predictive of classification correctness?\\
\textbf{RQ4:} Does training specialized models for specific constructs outperform a single generalist model in detecting misclassifications?

\section{Methodology}

The current study relies on a novel two-step method to explore improving annotations through classifying misclassifications. The first step, described in Section \ref{sec:LLMAnnotation} uses a typical LLM qualitative coding scheme in which LLMs are prompted to annotate a data set, in this case classroom transcripts with \textit{Talk Moves} derived from accountable talk theory. As part of this step, the LLMs are also prompted to provide reasoning for their classification. In the second step, described in Section \ref{sec:MisClass}, we trained machine learning models to detect misclassifications (false positives) using the reasoning output.

\subsection{Talk Moves Data Set}
We drew on the previously published \textit{TalkMoves Dataset}, which contains classroom transcripts annotated and validated by expert educators and is described in detail in prior work \cite{Suresh_2022_TalkMoves}. For the present study, we used a stratified sample of 63 transcripts covering multiple grade bands (sampling procedure described below). The corpus consists of mathematics instruction transcripts from authentic K–12 settings, including whole-class discussions, small-group work, and online lesson environments. Transcripts were human-transcribed, speaker-segmented, and labeled at the utterance level using a six-construct \textit{Talk Moves} scheme aligned with the Accountable Talk framework. Constructs associated with accountability to the learning community include \textit{Keeping Everyone Together}, \textit{Getting Students to Relate to Another’s Ideas}, and \textit{Restating}. Constructs associated with accountability to content knowledge include \textit{Pressing for Accuracy}. Constructs associated with accountability to rigorous thinking include \textit{Revoicing} and \textit{Pressing for Reasoning}. Utterances that did not exhibit any of these discourse moves were assigned a \textit{None} label.

The human-produced labels were treated as the reference standard for all performance comparisons. These labels were generated by trained mathematics educators using an established Accountable Talk coding protocol, with prior reporting showing very high inter-rater agreement (Cohen’s $\kappa > 0.90$) \cite{Suresh_2022_TalkMoves}. For model evaluation, we reorganized transcripts into context windows of up to 20 utterances, a preprocessing step designed to fit model context limits while maintaining local discourse continuity.

To manage computational and API cost, we conducted the study on a structured subsample rather than the full corpus. We applied proportional stratified random sampling to select 800 focal utterances so that the class distribution matched the original Talk Move label frequencies. Each focal utterance was then augmented with its surrounding dialogue (up to 20 preceding turns and one following turn) and overlapping windows were consolidated into continuous segments. Because context windows were included by design, the final set exceeded the initial focal count: about 79.5\% (10,522) were unique teacher utterances with valid Talk Move codes, and the remaining 20.5\% were additional turns retained to preserve conversational context. 

\subsection{Annotation Procedure}
\label{sec:LLMAnnotation}

We annotated the data sample for six talk moves constructs using four state-of-the-art Large Language Models (LLMs): \textbf{GPT-5} (OpenAI), \textbf{Claude 4.5 Sonnet} (Anthropic), \textbf{Gemini 2.5 Pro} (Google), \textbf{o3} (OpenAI). We employed a simple \textbf{zero-shot prompting strategy} with definitions of TalkMoves taken from the Talk Moves coding schema\footnote{\url{https://github.com/SumnerLab/TalkMoves/blob/main/Coding\%20Manual.pdf}}. The prompt included a request to ``provide a concise explanation in the 'Reasoning' field only when a Talk Move is present."  This generated a diverse dataset of LLM reasoning, ensuring the classifier was trained on a rich distribution of valid and invalid arguments across different generation attempts. This process generated an initial pool of over $30,817$ LLM predicted annotations. For 518 annotations ($1.8\%$), the models did not produce reasoning. This results in a total of $30,300$ annotations used in the analysis.

\subsection{Misconception Classification}
\label{sec:MisClass}
\subsubsection{Study Context \& Data Partitioning}

The dataset consists of {$N=30,300$} instances of teacher-student interactions, derived from a stratified sample of classroom logs. Each instance includes a target variable, \texttt{IsCorrect} (binary: 0 or 1), establishing ground truth for the correctness of the LLM's classification.
To ensure robust evaluation, we employed a stratified random sampling technique to partition the data into a {Training Set ($D_{train}$): 80\% of the corpus ($n=24,190$), used for model fitting and hyperparameter tuning; and a Test Set ($D_{test}$): 20\% of the corpus ($n=6,110$), held out strictly for final performance evaluation. 
Stratification was performed based on the \texttt{IsCorrect} label to preserve the class distribution (Correct vs. Incorrect) inherent in natural model processing.

\subsubsection{Feature Representation Details}
The primary input variable $X$ is the unstructured text of the reasoning for the annotation generated by the LLM. We transformed this raw text into a numerical feature space using Term Frequency-Inverse Document Frequency (TF-IDF) vectorization.  
The feature space dimension $V$ had a maximum of 3,227 unigrams ($|V|=3,227$), based on the max unigrams in the corpus. Stop-word removal (English) was applied to reduce feature noise. Preliminary experiments comparing Unigram vs. Bigram/Trigram architectures yielded negligible performance improvements ($<0.2\%$ $\Delta F1$), justifying the parsimonious choice of a unigram-only model.

\subsubsection{Classification Algorithms}
We implemented and compared five supervised classification algorithms to predict misclassification. All models incorporated class weighting (\texttt{class\_weight = 'balanced'}) to penalize misclassification of the minority class proportionally to its frequency.

\textbf{Logistic Regression (Baseline)}: Selected for its interpretation capabilities (coefficient analysis) and efficiency in high-dimensional sparse spaces.\\
\textbf{Random Forest (Ensemble)}: A bagging ensemble of 100 decision trees. This model was utilized to capture non-linear feature interactions (e.g., the conjunction of ``not'' and ``verbatim'') that linear models might miss.\\
\textbf{Support Vector Machine (Linear SVM)}: A max-margin classifier. We utilized a linear kernel to handle the high dimensionality of the TF-IDF space efficiently.\\
\textbf{Gradient Boosting (GBM)}: A boosting ensemble that builds simplified trees sequentially to minimize residual error, included to test robustness against hard-to-classify examples.\\
\textbf{Multinomial Naive Bayes}: A probabilistic baseline relying on Bayes' theorem with an assumption of feature independence, traditionally strong for text classification.

\subsection{Per-Construct Model}
To address the heterogeneity of linguistic correctness markers across different TalkMoves, we also ran one model per construct. This approach decomposes the global classification problem into a set of distinct sub-problems.
Let $C$ be the set of all possible TalkMove Constructs (e.g., ``Pressing for Reasoning'', ``Revoicing''). For each construct $c \in C$, we defined a specialist subset $D_c \subset D_{train}$ such that:
\begin{equation}
    D_c = \{(x_i, y_i) \mid \text{PredictedLabel}(x_i) = c\}
\end{equation}
For the top $k=6$ most frequent constructs, we trained an independent Random Forest classifier $M_c$ exclusively on $D_c$. At inference time, the system routes a new instance $x_{new}$ to the appropriate model $M_c$ based on its predicted label. This allows $M_c$ to learn a localized vocabulary of correctness—for instance, weighting the term ``classmate'' heavily for \textit{Getting Students to Relate}, a signal that might be irrelevant or noisy for \textit{Pressing for Accuracy}.

\begin{figure*}
    \centering
    \includegraphics[width=1\linewidth]{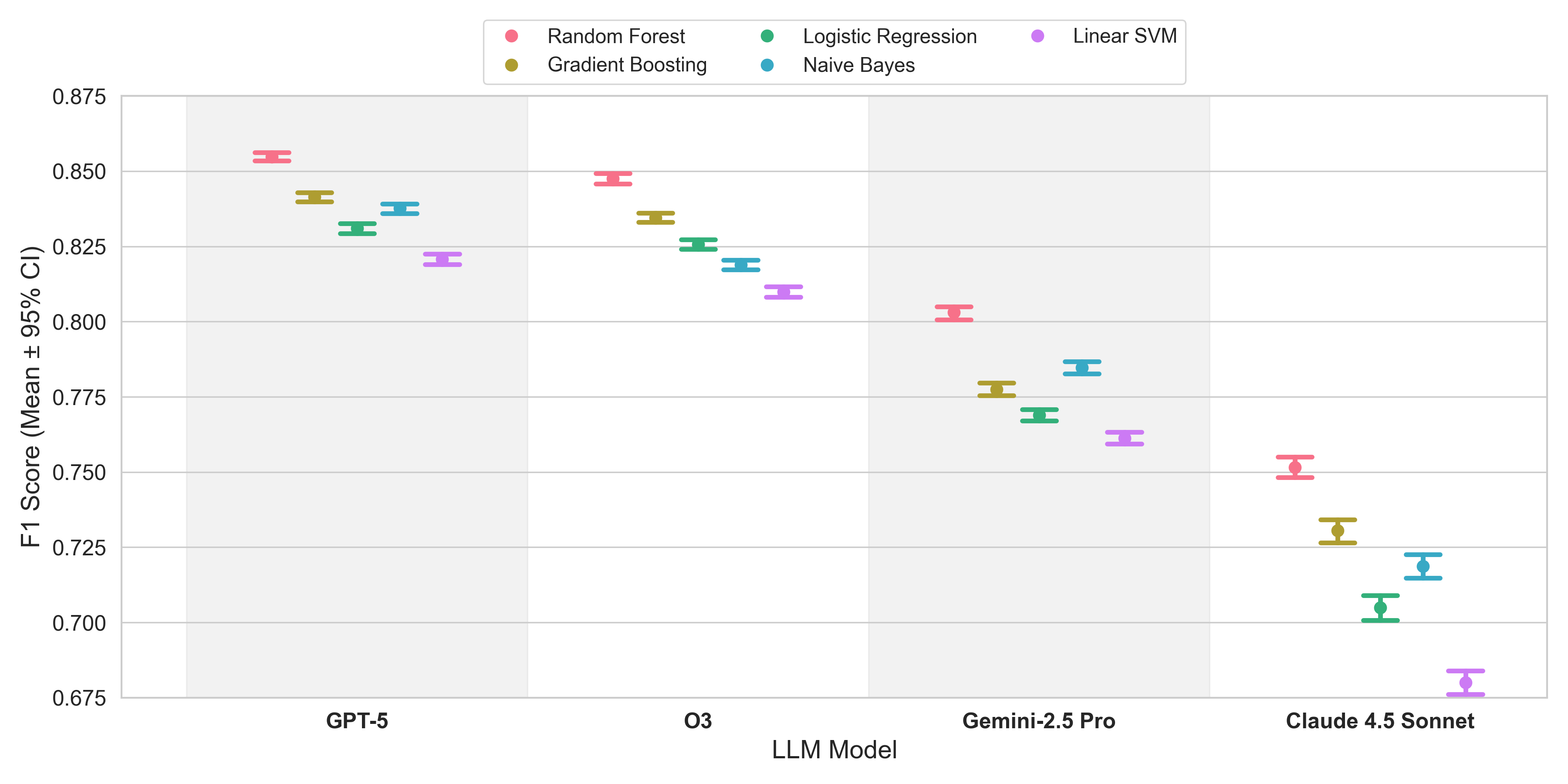}
    \caption{\textbf{Cross-Model Detection Performance.} A comparison of F1 scores for five supervised classification algorithms (Logistic Regression, Random Forest, SVM, Gradient Boosting, Naive Bayes) across the four target LLMs (Claude 4.5 Sonnet, Gemini 2.5 Pro, GPT-5, o3). The Random Forest classifier (pink points) consistently outperforms linear baselines across all generative models. Error bars represent 95\% confidence intervals.}
    \label{fig:performance}
\end{figure*}

\subsection{Linguistic Features (LIWC)}
To systematically analyze the cognitive and rhetorical structure of LLM reasoning, we processed all reasoning outputs using the Linguistic Inquiry and Word Count (LIWC) framework \cite{pennebaker2015development}. We focused on four theoretically relevant constructs: (a) logical grounding, (b) epistemic calibration, and (c) cognitive self-referential framing
We selected these linguistic feature constructs to operationalize dimensions of reasoning quality and epistemic stance in LLM-generated explanations. Prior work on model explanations and uncertainty suggests that incorrect or weakly grounded reasoning is often characterized not by syntactic simplicity, but by epistemic hedging, metacognitive narration, and reduced evidential linkage between claims and conclusions \cite{chen2022explaining, ji2023survey}. 

\textbf{Causation} markers (e.g., \emph{because, therefore, implies}) were included to measure explicit premise–conclusion linking. In instructional move classification, correct labels require mapping observable discourse evidence to construct definitions. Reasoning that explicitly encode causal or justificatory structure may be more likely to reflect evidence-based classification rather than keyword matching or surface heuristics \cite{biran2017human}.

\textbf{Differentiation} markers (contrastives, discrepancies, and exclusions; e.g., \emph{but, however, except, should}) were included to measure contrastive and error-checking structure. Many construct definitions involve ruling out alternative interpretations (e.g., distinguishing revoicing from restating). We therefore tested whether valid reasoning exhibit greater contrastive structure. This composite category captures logical branching and boundary-setting language.

\textbf{Tentativeness} markers (e.g., \emph{might, could, possibly}) were selected to capture epistemic hedging and uncertainty expression. Work on verbalized model uncertainty shows that LLMs frequently externalize low confidence through modal and hedge language \cite{ji2023survey}. We therefore treat tentativeness density as a proxy for internal uncertainty signals that may correlate with misclassification risk \cite{chen2022explaining}.

\textbf{Insight} markers (e.g., \emph{think, realize, know}) were included to detect metacognitive and self-referential reasoning frames.  For our proposal, these marks might be indicative of stances rather than empirically grounded evidence. Prior work on stance and epistemic modality shows that verbs such as “think” and “know” are linguistic markers of subjective commitment or metacognitive framing rather than text-based evidential support, and have been used as features in uncertainty/stance tasks in NLP \cite{de2018stance}.

Together, these feature groups map onto complementary dimensions for exploratory analysis of LLM reasoning and its association with misclassification. This design enables direct testing of whether correctness is primarily associated with logical grounding versus rhetorical or metacognitive elaboration. Finally we examine the differences in reasoning length between correct and incorrect annotations. A common heuristic in prompt engineering is that ``more reasoning equals better performance'' \cite{wei2022chain}. However, this is often used in the context of `chain-of-thought' reasoning. Less is known about how explanation length behaves when models are prompted to justify a label they have already produced. Accordingly, we analyze the distribution of rationale lengths for correct and incorrect annotations to assess whether explanation length provides a useful signal for verification.

\section{Results}

\subsection{Can LLM Reasoning Predict Correctness? (RQ1)}
\label{sec:results_classification}

First, we evaluate whether the LLM reasoning text contains sufficient signal to verify the correctness of an LLM's label by benchmarking classifier performance on the holdout sample. As shown in Table \ref{tab:classifier_benchmarks}, all models achieved high predictive accuracy, with $F_1$ scores ranging from 0.79 to 0.83. The Random Forest classifier yielded the top performance, achieving a Precision of 0.807 and a Recall of 0.854. This high recall is particularly critical in an educational context, as it minimizes the risk of inadvertently discarding valid Talk Moves during the verification process. Overall, these results provide strong evidence that LLM reasoning contains latent linguistic markers that effectively signal classification accuracy.
\begin{table}[h]
    \centering
    \caption{\textbf{Performance of Verification Classifiers.} Evaluation on the held-out test set ($N=6,110$). Random Forest yields the highest F1 score.}
    \resizebox{\columnwidth}{!}{
        \begin{tabular}{lcccc}
            \hline
            \textbf{Model} & \textbf{Precision} & \textbf{Recall} & \textbf{F1} & \textbf{Accuracy} \\
            \hline
            Random Forest & \textbf{0.807} & 0.854 & \textbf{0.830} ($\pm$ 0.009) & \textbf{81.1\%} \\
            Gradient Boosting & 0.764 & \textbf{0.867} & 0.812 ($\pm$ 0.010) & 78.3\% \\
            Naive Bayes & 0.783 & 0.832 & 0.807 ($\pm$ 0.010) & 78.5\% \\
            Logistic Regression & 0.806 & 0.797 & 0.802 ($\pm$ 0.010) & 78.7\% \\
            Linear SVM & 0.798 & 0.780 & 0.789 ($\pm$ 0.011) & 77.4\% \\
            \hline
        \end{tabular}
    }
    \label{tab:classifier_benchmarks}
\end{table}

To contextualize this prediction task, consider the following two examples of model-generated reasoning drawn from our dataset, only one of which is correct. 

\begin{quote}
\textit{``The teacher rephrases the student's question, clarifying the idea that the measurement might not be the whole volume because the object is floating.''} 
\hfill --- \textbf{Reasoning A} (Gemini 2.5 Pro)
\end{quote}
\begin{quote}
\textit{``The teacher explicitly asks the student to provide the reasoning ('why') for their answer.''} \\
\hfill --- \textbf{Reasoning B} (Gemini 2.5 Pro)
\end{quote}

Recognizing that Reasoning A accompanied a misclassification, while Reasoning B presents a correct justification is not obvious. Notice that the incorrect reasoning leaks uncertainty through hedging or speculative language ("might not be"), whereas the correct reasoning is more direct and grounded in observable instructional actions ("explicitly asks"). We formally analyze linguistic markers in Section 5.3.

\subsection{Robustness Across LLMs (RQ2)}
We evaluated classifier performance across the four distinct LLMs used for annotation: GPT-5, o3, Gemini 2.5 Pro, and Claude 4.5 Sonnet. As illustrated in Figure \ref{fig:performance}, all misconception classification models performed well ($F1>0.67$) on classifications from all LLMs. The Random Forest classifier consistently outperformed the other models across all four LLMs. This dominance was particularly pronounced for annotations generated by Claude 4.5 Sonnet, where other models dropped to $F1$ scores as low as $0.70$, while the Random Forest classifier maintained a significantly higher performance of $0.75$. This performance gap indicates that the linguistic features distinguishing valid from invalid reasoning may be inherently nonlinear, necessitating the complex decision boundaries of ensemble methods regardless of the source LLM.

While the detection method remained effective globally, the absolute detectability of misclassifications varied by source model. Misclassifications generated by reasoning models like o3 and GPT-5 were the easiest to detect, with the Random Forest achieving $F1$ scores exceeding $0.85$. The strong agreement between linear and nonlinear models in these conditions suggests that errors in these models may be marked by distinct, easily separable lexical artifacts. In contrast, misclassifications from Claude 4.5 Sonnet proved more challenging to verify. In these cases, the generated reasoning may be more linguistically homogenous with valid reasoning, reducing the effectiveness of simple keyword-based detection. Overall, while the approach performs moderately well across the board, performance is maximized when applied to models that exhibit distinct and detectable misclassifications.
\begin{figure}[h]
    \centering
    \includegraphics[width=1\linewidth]{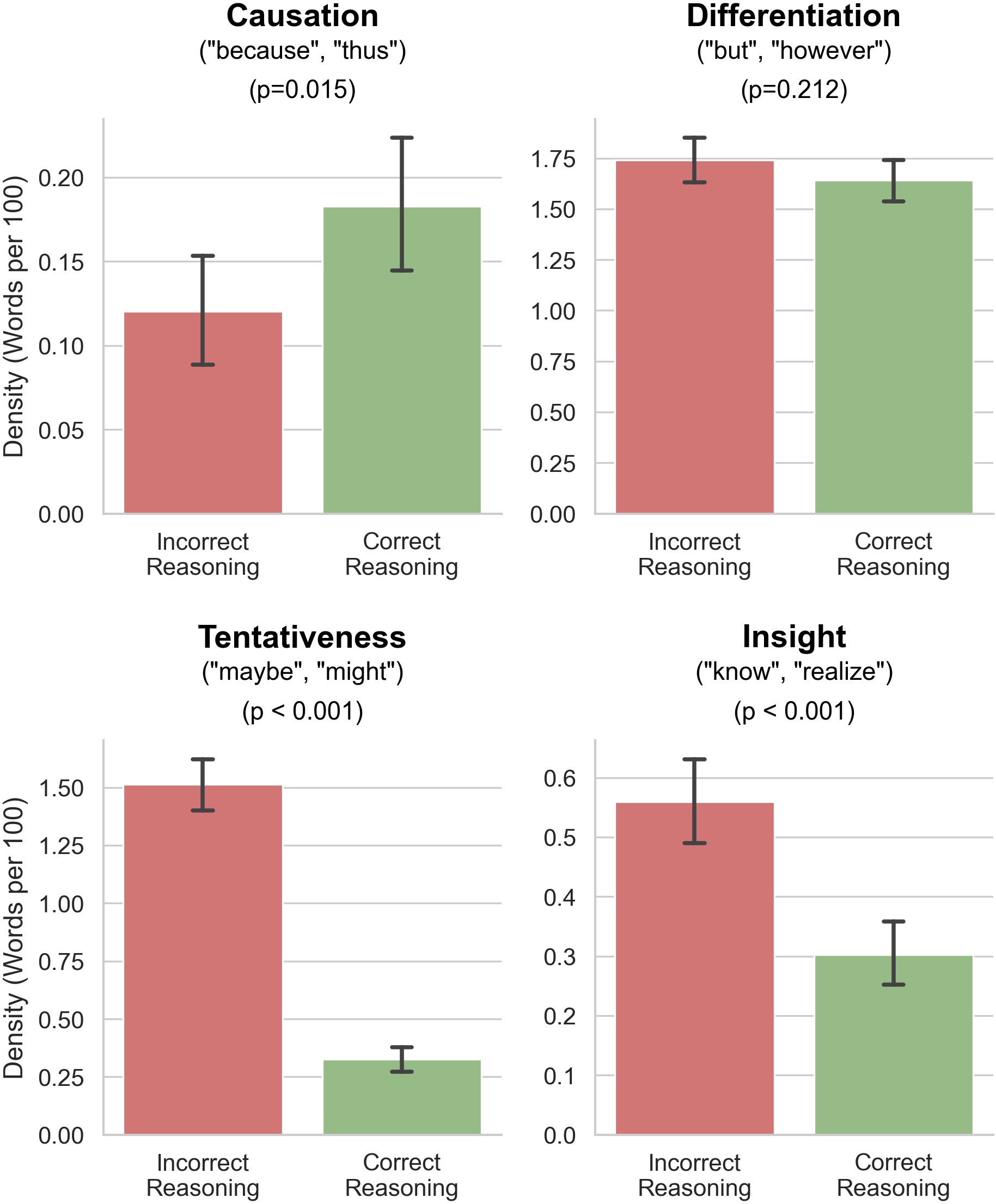}
    \caption{Linguistic Markers of Correct vs. Incorrect Reasoning. Average density of linguistic features (words per 100 words) in correct (orange) vs. incorrect (blue) mathematical explanations. Causation (``therefore", ``because") is the primary predictor of correctness ($p=0.015$), indicating that explicit logical structure aligns with accurate problem solving. Conversely, Tentativeness (``might", ``could") and Insight (``believe", ``think") are more prevalent in incorrect responses ($p<0.001$), suggesting that models use hedging and metacognitive language as ``filler" when reasoning is flawed or uncertain. Differentiation (``but", ``however") showed no significant difference ($p=0.212$), appearing equally in both valid and invalid arguments. Error bars represent 95\% confidence intervals.}
    \label{fig:linguistic}
\end{figure}
        
\subsection{Linguistic Markers of Misclassification (RQ3)}
\label{sec:linguistic_signatures}
Our multidimensional linguistic analysis reveals distinct cognitive profiles for valid versus invalid reasoning. As illustrated in Figure \ref{fig:linguistic}, correctness in mathematical reasoning is not signaled by complexity or "insight," but by the prosaic density of causal logic.

\subsubsection{Metacognition as a Negative Signal}
Insight was a significant \textit{negative} signal ($p < 0.001$). Reasoning that contained high densities of metacognitive terms like "I think," "I realize," and "understand" were overwhelmingly more likely to be incorrect. This finding identifies a specific mode of failure: mere "performative reasoning." When the model lacks a clear logical path, it compensates by narrating its own cognitive state, filling the reasoning with vacuous self-reflection instead of derivation. 

\subsubsection{Verbalized Uncertainty and Calibration}
We examined \textit{Tentativeness} (e.g., ``might,'' ``could,'' ``possibly'') to determine if models effectively signaled their uncertainty and found a statistically significant trend where incorrect answers contained a higher density of tentative language. This indicates a degree of successful calibration: when the model is less certain of its reasoning, it employs more hedging. Conversely, correct reasoning was characterized by declarative, confident assertions, suggesting that high certainty (low tentativeness) is a reliable linguistic proxy for mathematical accuracy. This "Verbalized Uncertainty" \cite{lin2022teaching} serves as a potent diagnostic signal, allowing the classifier to detect when the model is "guessing" based on weak correlations rather than identifying a clear signal.

\subsubsection{Causal Chaining and Logical Validity}
Valid mathematical derivations are fundamentally characterized by explicit causal structures. Likewise, we found that correct responses exhibited a significantly higher ($p=0.015$) density of {Causation} markers (e.g., ``therefore,'' ``because,'' ``implies,'' ``yields'') than misclassified responses. Correct reasoning contained a mean density of 0.18 causal markers per 100 words, compared to only \textit{0.12} for misclassified responses. This finding confirms that accuracy is strongly correlated with the use of directional logical connectors: correct models explicitly link premises to conclusions, whereas incorrect models often present disconnected assertions without the requisite connective tissue.

\subsubsection{Conceptual Differentiation and Structural Complexity}
We analyzed the density of Differentiation markers (e.g., ``but,'' ``however,'' ``except'') to determine if valid reasoning involved more robust error-checking or case handling. The analysis yielded a null result ($p=0.212$), with no significant difference in differentiation density between correct ($\mu=1.63$) and incorrect ($\mu=1.74$) reasoning. This indicates that incorrect reasoning is often just as syntactically complex as valid reasoning. Reasoning of misclassified responses frequently employs contrastive language not to distinguish valid mathematical constraints, but rather to express confusion or hedge between incorrect options, resulting in a linguistically sophisticated but logically flawed explanation.
\begin{figure}[t]
    \centering
    \includegraphics[width=1\linewidth]{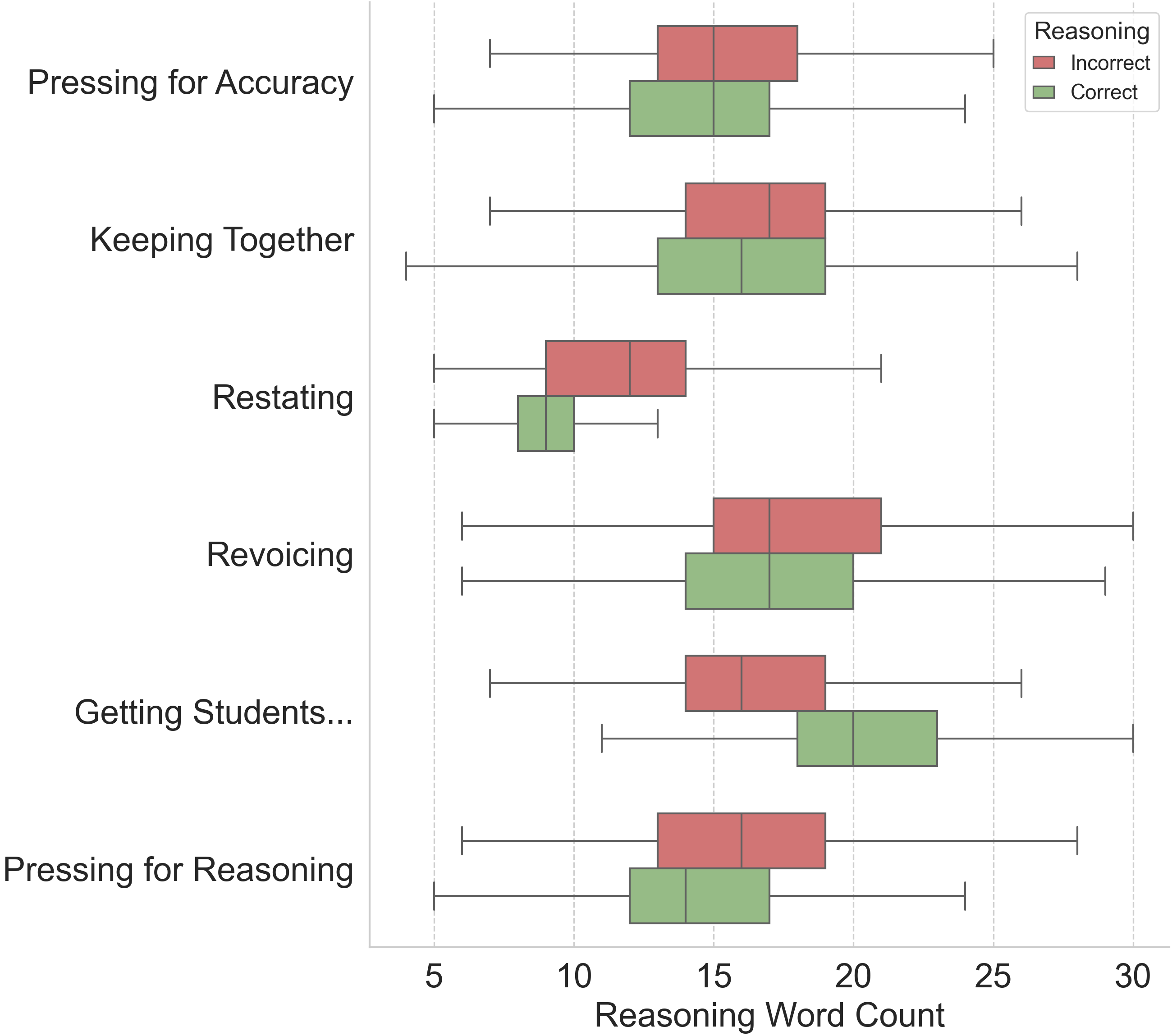}
    \caption{{\textbf{Reasoning Length Distribution.} Distribution of reasoning lengths for Correct (Green) vs. Incorrect (Red) predictions. All differences are statistically significant ($p < 0.05$). Reasoning for correct responses is generally more concise than for incorrect responses, with the notable exception of \textit{Getting Students to Relate}, where correct identification of complex student-to-student interactions requires significantly longer explanations.}}
    \label{fig:length}
\end{figure}

\subsubsection{Reasoning Length as a Signal}
Besides the LWIC analysis, we also evaluated whether the reasoning length analysis differed based on correct or incorrect responses. Figure \ref{fig:length} reveals distinct verbosity profiles for valid vs. invalid predictions: For the majority of constructs (e.g., \textit{Pressing for Reasoning}, \textit{Keeping Everyone Together}), valid reasoning (Green boxplots) is \textbf{significantly shorter} than incorrect ones (Red boxplots). This suggests that valid reasoning is precise, evidence-based, and directed. In contrast, misclassified reasoning tends to be verbose and circuitous, often characterized by the model "waffling" or attempting to rationalize a weak connection with filler text.
The "Relating" Exception: The sole statistically significant exception is \textit{Relating to Others}, where correct reasoning is longer ($Median \approx 22$ words) than misclassifications. This makes intuitive sense: correctly identifying a "Relate" move requires describing a complex interaction between the teacher, Student A, and Student B. A misclassification in this construct often fails to capture this tripartite structure, resulting in shorter, more vague reasoning.
\begin{figure}[t]
    \centering
    \includegraphics[width=1\linewidth]{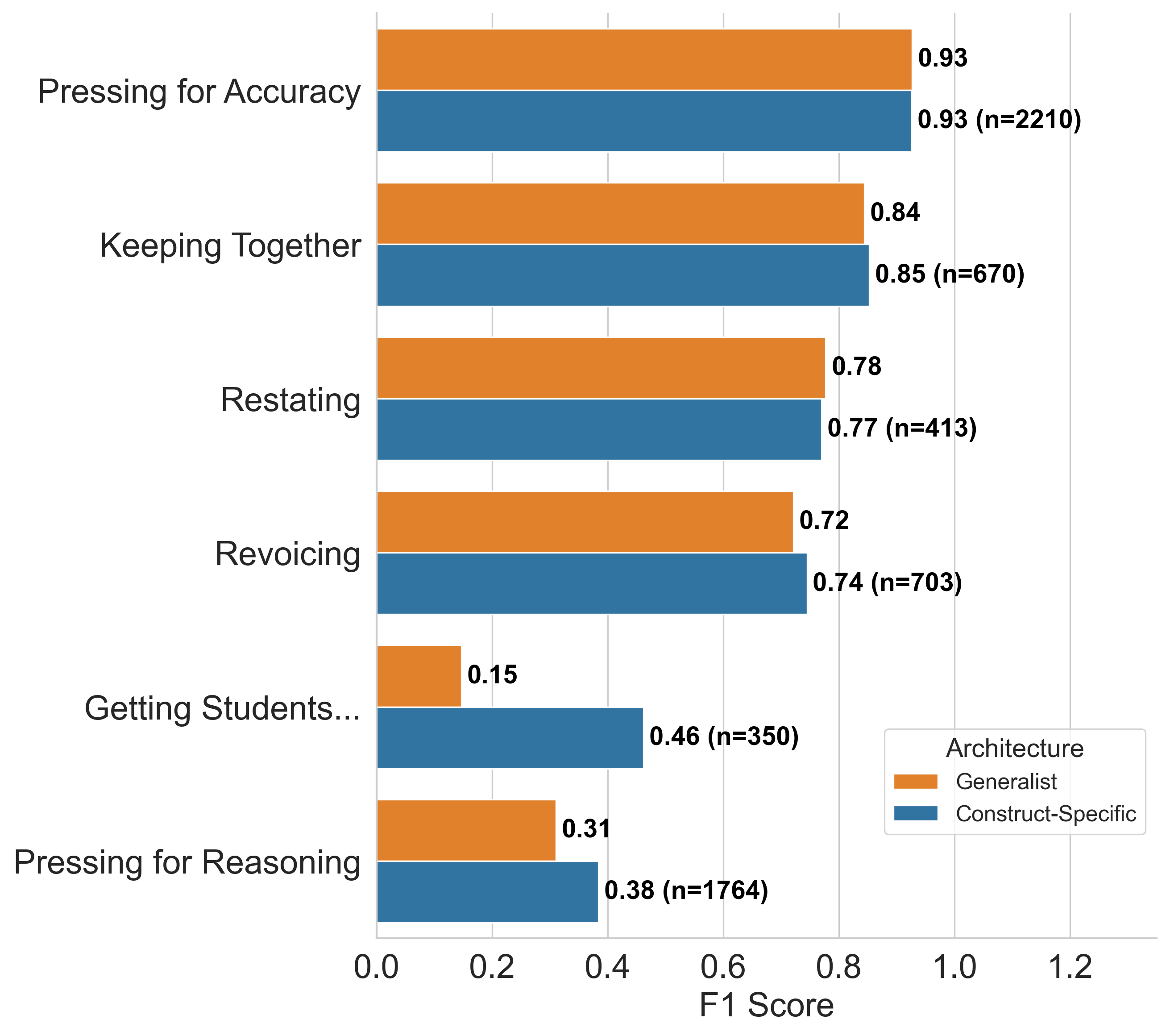}
    \caption{\textbf{Performance Comparison: Generalist vs. Construct-Specific Models.} While generalist models perform comparably on constructs like \textit{Pressing for Accuracy} ($F1 \approx 0.93$), construct-specific models provide critical gains for complex moves. Notably, for \textit{Getting Students to Relate} ($n=350$), the specialist architecture triples the F1 score (0.46 vs. 0.15), recovering signal that is otherwise drowned out by global patterns.}
    \label{fig:comparison}
\end{figure}

\subsection{Construct-Specific Detection (RQ4)}
\label{sec:specialist_arch}
Finally, we investigated whether we could optimize performance by training Construct-Specific Verifiers—dedicated Random Forest models trained exclusively on partitions of the dataset corresponding to a single TalkMove. As illustrated in Figure \ref{fig:comparison}, these targeted models demonstrated distinct performance profiles based on construct complexity. 
We observed high verification reliability for \textbf{Pressing for Accuracy} ($F1 = 0.86$) and \textbf{Restating} ($F1=0.78$). This indicates that for moves defined by specific keywords (e.g., "exact words", "vocabulary"), the classifier is effective at distinguishing valid reasoning from misclassifications.
In contrast, inferential constructs proved more challenging. \textbf{Pressing for Reasoning} achieved an F1 of 0.48. Similarly, \textbf{Getting Students to Relate to Another's Ideas} ($n=1,918$) achieved an F1 of 0.46. While low in absolute terms, this represents a significant signal recovery compared to the generalist baseline ($F1=0.15$), confirming that training on a dedicated subset captures distinct tripartite references (Teacher-Student-Student) that are otherwise diluted in the global dataset.

\begin{table*}[h!]
    \centering
    \caption{\textbf{Linguistic Markers of Correctness (TF-IDF Analysis).} A breakdown of the lexical markers that distinguish correct reasoning from misclassifications across six TalkMoves. (+) denotes positive predictors, (-) denotes negative predictors.}
    \resizebox{\linewidth}{!}{
    \begin{tabular}{p{0.20\linewidth} p{0.30\linewidth} p{0.23\linewidth} p{0.23\linewidth}}
        \toprule
        \textbf{Construct} & \textbf{Key Markers} & \textbf{Correct Example (+)} & \textbf{Incorrect Example (-)} \\
        \midrule
        \textbf{Pressing for Reasoning} & (+) \textit{justify, because, explain why} \newline (-) \textit{share, thinking, thoughts} & "The teacher asks 'How do you know', prompting the student to explain their \textbf{justification}." & "The teacher prompts students to write... eliciting explanation of their \textbf{thinking}." \\
        \cmidrule{1-4}
        \textbf{Restating} & (+) \textit{verbatim, repeats, exactly} \newline (-) \textit{clarify, rephrase, confirm} & "The teacher repeats the student's response ('No') \textbf{verbatim}." & "The teacher repeats a key part... for \textbf{clarification}, changing the words." \\
        \cmidrule{1-4}
        \textbf{Revoicing} & (+) \textit{paraphrases, repeats, wording} \newline (-) \textit{question, check, understanding} & "The teacher \textbf{paraphrases} the student's idea... changing \textbf{wording} to clarify." & "The teacher reframes... by suggesting 'dis' might mean 'wrong,' checking \textbf{understanding}." \\
        \cmidrule{1-4}
        \textbf{Pressing for Accuracy} & (+) \textit{precise, specific, term} \newline (-) \textit{check, confirm, correctness} & "The teacher asks the student to use the \textbf{precise} mathematical \textbf{term}." & "The teacher \textbf{checks} whether a specific way of writing is valid..." \\
        \cmidrule{1-4}
        \textbf{Keeping Everyone Together} & (+) \textit{class, shared, orienting} \newline (-) \textit{agreement, explain, question} & "Prompts a student to \textbf{read aloud} to the \textbf{class}, orienting listeners." & "The teacher asks a \textbf{question} to ensure all students are following." \\
        \cmidrule{1-4}
        \textbf{Getting Students to Relate} & (+) \textit{classmate, peer, react} \newline (-) \textit{connect (vague), link, interaction} & "The teacher explicitly asks students to \textbf{react} to [Student]'s idea." & "The teacher asks students to \textbf{connect} their answer to the previous problem." \\
        \bottomrule
    \end{tabular}
    }
    \label{tab:linguistic_markers}
\end{table*}

\subsubsection{Linguistic Markers of Correctness: A Construct-by-Construct Analysis}
To understand \textit{what} the specialist classifiers are learning beyond global heuristics, we analyzed feature importance scores to extract the ``vocabulary of correctness'' for each TalkMove. Table \ref{tab:linguistic_markers} details the distinguishing signatures and provides in-context examples of correct vs. incorrect reasoning.

This analysis reveals that reasoning for each pedagogical move has a \textit{linguistic fingerprint}:

\textbf{Explicit Moves (e.g., Restating):} Reliability is driven by correct reasoning containing synonyms for exactness (\textit{verbatim, repeats, word-for-word}). Misclassifications typically involve words related to modification (\textit{clarify, rephrase}), indicating the model is confusing a repetition with a re-explanation.

\textbf{Inferential Moves (e.g., Pressing for Reasoning):} The signal relies on causal language (\textit{because, justify, why}). Misclassifications often use vague cognitive terms (\textit{thinking, thoughts}), failing to distinguish between a general question and a specific request for evidence.

\textbf{Social Moves (e.g., Relating, Keeping Together):} Correctness is marked by collective nouns (\textit{class, peer, classmate, we}). The absence of these group-orienting terms is a strong predictor of misclassifications, where the model falsely attributes a social move to a generic teacher-student interaction.

\section{Discussion}
This study shows that model-generated rationales (i.e., elicited explanations produced alongside automated discourse labels) contain systematic linguistic signals that predict whether an LLM’s annotation is correct. Across various LLMs, prediction models, constructs, and verification architectures, we find that these rationales support reliable detection of misclassifications (F1 $\approx$ 0.83), demonstrating that error and uncertainty are frequently externalized when models are prompted to explain their decisions. This finding aligns with prior work showing that reasoning and verbalized confidence signals encode information about model uncertainty and correctness \cite{kadavath2022language,tian2023just,xiong2025trace}. Rather than treating explanations as auxiliary artifacts, our results position them as a practical verification layer that can be leveraged to improve the reliability of large-scale automated coding in EDM, where LLM-based classification is increasingly used for complex, high-inference constructs \cite{kim2025code,neshaei2025bridging}.

These findings contribute to ongoing debates about the role and reliability of LLM explanations. Prior work has cautioned that explanations may function as post-hoc rationalizations that are not causally responsible for model predictions \cite{turpin2023language,ji2023survey,agarwal2024faithfulness}. Our results support a more application-oriented interpretation: even when elicited rationales are not guaranteed to be faithful to internal model mechanisms, their linguistic structure remains predictively diagnostic of correctness. In high-inference educational coding tasks where reliability has been shown to vary widely across prompts and constructs \cite{zarisheva2024deductive,vanacore2025well}, valid annotations are more often supported by rationales grounded in observable textual evidence and explicit causal logic, whereas incorrect annotations more frequently rely on speculative or self-referential language. This enables downstream verification without requiring access to model internals or computationally expensive multi-sample strategies such as self-consistency \cite{wang2022self} or SelfCheck-style approaches \cite{manakul2023selfcheckgpt}.

The linguistic analysis reveals a consistent pattern across models and constructs. Correct annotations are associated with concise, declarative rationales that explicitly connect utterance features to constructs using causal connectors and evidence-grounded justification. In contrast, misclassifications are marked by increased epistemic hedging, performative metacognitive language, and greater verbosity. This pattern is consistent with prior analyses showing that LLM uncertainty and error are often expressed through longer, more hedged reasoning traces \cite{xiong2025trace} and with broader accounts of confabulation and faithfulness failures in generated explanations \cite{ji2023survey,agarwal2024faithfulness,yao2024editing}. LIWC-based and related linguistic profiling work further suggests that AI-generated text exhibits systematic shifts in hedging, insight, and causation markers when semantic grounding is weak \cite{pennebaker2015development,yao2024editing}. Importantly, incorrect rationales in our data are not linguistically simplistic; they are often syntactically complex and rhetorically fluent, reinforcing that surface sophistication alone is not a reliable proxy for validity.

We also observe substantial variation in detectability across LLM architectures. Misclassifications generated by GPT-5 and o3 were comparatively easy to identify, while errors from Claude~4.5~Sonnet were more difficult to distinguish from correct annotations. This result is consistent with prior findings that LLM annotation reliability varies significantly across models and prompt formulations \cite{kim2025code,zarisheva2024deductive}. It also suggests that as models improve in rhetorical fluency, incorrect rationales increasingly approximate the discourse patterns of correct ones, creating a moving-target problem for explanation-based verification. Explanation-based detection remains effective, but its discriminative margin may narrow as stylistic quality improves. This reinforces the need to combine rationale-based verification with complementary approaches such as calibration signals and construct-specific prompting strategies \cite{kadavath2022language,tian2023just}.

Our results further highlight a trade-off between model generality and construct specificity in LLM-based coding. Prior work shows that LLM performance drops for ambiguous or high-inference constructs and that targeted prompt and example design can substantially improve results \cite{tran2025collaborative,petrilli2024next,kim2025code}. Consistent with this pattern, we find that specialist verifiers yield substantial gains for certain discourse constructs, suggesting a practical pipeline in which general-purpose LLM annotation is paired with construct-targeted verification layers. Linguistic profiles of correct rationales—such as explicit relational or causal structure—provide actionable features for designing these construct-specific verifiers and prompts.

Finally, these results have direct implications for educational data mining pipelines that rely on automated discourse annotation. LLM-based coding is increasingly used to analyze tutoring dialogue, classroom discourse, and other forms of unstructured educational text \cite{acosta2025recognizing,zhang2024detecting,tran2024analyzing,whitehill2023automated}. However, prior studies document persistent false positives and construct-level reliability gaps \cite{vanacore2025well,zarisheva2024deductive}. Because downstream methods such as discourse modeling and instructional quality measurement are sensitive to systematic labeling errors, explanation-based verification offers a scalable quality-control layer. By using rationale-derived signals to triage outputs, high-confidence annotations can be accepted automatically while flagged cases are routed for human review or construct-specific reanalysis. This selective acceptance strategy supports scalable annotation while preserving inferential validity, advancing more trustworthy deployment of LLMs for educational discourse analysis.

\subsection{Limitations}
Our study focused on a specific math instructional context. While the concept of detection via reasoning is likely generalizable, the specific vocabulary (e.g., ``numerator'', ``fraction'') may be domain-specific. Future work should test the transferability of these Correctness markers to other domains, such as English Language Arts or Science. All rationales analyzed in this study were elicited through a specific prompting strategy that explicitly asked models to justify their labels. Different prompt formulations, levels of instruction, or constraints on explanation length may influence how uncertainty and error are linguistically expressed. As a result, the strength and form of the verification signal may vary under alternative prompting regimes. Our models operate on textual features of model-generated rationales using traditional NLP representations. While this design choice improves interpretability and computational efficiency, it does not capture deeper semantic or discourse-level structure that may further distinguish valid from invalid explanations, particularly for high-inference constructs. More expressive representations may improve detection performance but also introduce additional complexity and cost.

\section{Conclusion}
As LLMs become increasingly integrated into educational data mining workflows, the reliability of automated annotation remains a central challenge. This work shows that when models are prompted to justify their labels, the resulting model-generated rationales contain systematic linguistic signals that can be used to detect misclassifications at scale. Across multiple state-of-the-art LLMs, instructional constructs, and verification architectures, we demonstrate that these rationales support accurate prediction of annotation correctness, offering a practical mechanism for quality control in automated discourse analysis.

Beyond performance gains, our findings provide insight into how and when automated annotations fail. Correct annotations are supported by concise, causally grounded explanations that reference observable features of the discourse, while incorrect annotations are more often accompanied by hedging, performative metacognitive language, and increased verbosity. These patterns suggest that elicited explanations externalize model uncertainty in a way that is both systematic and exploitable, even when explanations are not guaranteed to reflect internal decision processes.

From an applied perspective, this work reframes explanations from passive artifacts into actionable verification signals. Rather than treating LLM outputs as binary predictions, educational researchers can leverage model-generated rationales to selectively accept high-confidence annotations and flag uncertain cases for further review. This approach enables scalable annotation while mitigating the propagation of false positives into downstream analyses such as sequential modeling, instructional quality measurement, and network-based representations of classroom discourse.

More broadly, this study contributes to ongoing discussions about the role of explanations in machine learning systems. While explanations may not always be faithful representations of internal reasoning, our results show that they can nonetheless be predictively useful for assessing correctness in real-world, high-inference tasks. By operationalizing explanation-based verification in an educational setting, this work offers a practical path toward building more trustworthy, transparent, and robust LLM-powered analytics for learning and instruction.

%\end{document}  % This is where a 'short' article might terminate

%ACKNOWLEDGMENTS are optional
% \section{Acknowledgments}
% This section is optional; it is a location for you
% to acknowledge grants, funding, editing assistance and
% what have you.  In the present case, for example, the
% authors would like to thank Gerald Murray of ACM for
% his help in codifying this \textit{Author's Guide}
% and the \textbf{.cls} and \textbf{.tex} files that it describes. Acknowledgments should be left blank during the review process.

\bibliographystyle{abbrv}
\bibliography{manuscript} %sigproc.bib is the name of the Bibliography in this case
\end{document}